\documentclass[12pt]{elsarticle}
\usepackage{graphicx}
\graphicspath{{./images/}}
\usepackage{url}

\usepackage{amssymb}

\begin{document}

\begin{frontmatter}



\title{Improving Children’s Speech Recognition by Fine-tuning Self-supervised Adult Speech Representations}


\author[inst1]{Renee Yonghan Lu}

\affiliation[inst1]{organization={School of Electrical and Computer Engineering},
            addressline={University of New South Wales}, 
            city={Sydney},
            postcode={2052}, 
            state={NSW},
            country={Australia}}

\author[inst1]{Mostafa Shahin}
\author[inst1]{Beena Ahmed}

\begin{abstract}
Children’s speech recognition is a vital, yet largely overlooked domain when building inclusive speech technologies. The major challenge impeding progress in this domain is the lack of adequate child speech corpora; however, recent advances in self-supervised learning have created a new opportunity for overcoming this problem of data scarcity. In this paper, we leverage self-supervised adult speech representations and use three well-known child speech corpora to build models for children’s speech recognition. We assess the performance of fine-tuning on both native and non-native children’s speech, examine the effect of cross-domain child corpora, and investigate the minimum amount of child speech required to fine-tune a model which outperforms a state-of-the-art adult model. We also analyze speech recognition performance across children’s ages. Our results demonstrate that fine-tuning with cross-domain child corpora leads to relative improvements of up to 46.08\% and 45.53\% for native and non-native child speech respectively, and absolute improvements of 14.70\% and 31.10\%. We also show that with as little as 5 hours of transcribed children’s speech, it is possible to fine-tune a children’s speech recognition system that outperforms a state-of-the-art adult model fine-tuned on 960 hours of adult speech.
\end{abstract}


\end{frontmatter}


\section{Introduction}
\label{sec:intro}
Automatic speech recognition (ASR) systems have enjoyed many developments and innovations over the last decade and consequently rooted their presence in consumer technology. Although this has led to a growing demand for inclusive and broadly accessible speech technologies, sadly progress in children’s ASR has remained largely stagnant when compared with the rapid advancements that have occurred in the more conventional realm of adult speech recognition. This is despite the tremendous opportunities and benefits in child-centric speech technologies such as remote speech therapy tools \cite{hair2018apraxia}, interactive reading tutors \cite{mostow1994prototype}, pronunciation coaching \cite{russell1996applications}, and educational games.

The scarcity of transcribed child speech corpora combined with acoustic and linguistic differences between child and adult speech makes children’s speech a particularly challenging domain. Less than 20 child speech corpora exist worldwide, and only three of them are sufficiently large for training ASR systems \cite{ahmed2021auskidtalk}. It is because of this that the domain of children’s speech is especially suited to self-supervised speech representation learning methods, which extract high-level properties of speech by training networks with vast amounts of unlabeled speech data \cite{chung2019unsupervised}.

The two major techniques for self-supervised learning are predictive coding methods and transformer methods. Seminal works in predictive coding include Contrastive Predictive Coding (CPC) \cite{oord2018representation} and Autoregressive Predictive Coding (APC) \cite{chung2019unsupervised}. Extensions to these methods include the Deep Contextualized Acoustic Representations (DeCoAR) model \cite{ling2020deep} and vector-quantized APC model \cite{chung2020vector}.

More recently, transformer neural networks \cite{vaswani2017attention} have been applied in several works with wav2vec 2.0 \cite{baevski2020wav2vec} being one of the most well-known state-of-the-art architectures. The authors of \cite{baevski2020wav2vec} released several wav2vec 2.0 speech representation models that are pre-trained on enormous amounts of unlabeled adult speech, which can be fine-tuned for use in downstream tasks. The wav2vec 2.0 model uses a contrastive loss function to learn speech representations in a self-supervised manner by identifying the real future latent representation from false future representations. Other transformer-based self-supervised architectures include Mockingjay \cite{liu2020mockingjay}, Speech-XLNet \cite{song2019speech}, Audio ALBERT \cite{chi2021audio} and TERA \cite{liu2021tera}.

The current interest in self-supervised learning coupled with the popularity of wav2vec 2.0 has attracted a considerable amount of work in adapting pre-trained self-supervised models for downstream speech tasks in data scarce domains. In \cite{wu2021transformer}, the authors explored mispronunciation detection for non-native adult speakers, using data from Cantonese and Putonghua speakers to fine-tune a wav2vec 2.0 model that was pre-trained on speech from native English-speaking adults. Similarly, \cite{xu2021explore} used the same pre-trained wav2vec 2.0 model to build a non-native mispronunciation detection system. The work in \cite{shibano2021speech} used a pre-trained wav2vec 2.0 model to fine-tune a non-native adult ASR system, comparing single-accent training and multi-accent training. However, these utilized the same speech corpus for both single and multi-accent training, with no work done in fine-tuning with a cross-domain corpus consisting of speech from different domains. In \cite{deng2021improving}, the authors proposed an accent identification model to improve non-native adult ASR when fine-tuning a wav2vec 2.0 model.

Unfortunately, despite the wide range of work on self-supervised speech representation learning for adults, there is a dearth of literature on the usage of these methods for child speech. To the best of our knowledge, there have been only two studies on the use of self-supervised speech representation learning for children’s ASR: \cite{fan2021bi} explored a bidirectional-APC for native English-language children’s ASR and \cite{xu2021tal} explored wav2vec 2.0 for recognizing a small corpus of non-native English and German speech.

Our paper addresses this gap by assessing the performance of a state-of-the-art self-supervised speech representation architecture when used for children’s speech recognition to answer three key questions:
\begin{enumerate}
    \item What is the effect of fine-tuning: Can we build acceptable speech recognition models for native and non-native English-speaking children by fine-tuning a pre-trained self-supervised adult speech representation model?
    \item What is the effect of cross-domain corpora: Can we improve children’s ASR performance using cross-domain speech corpora?
    \item What is the effect of data quantity: How much children’s data do we need to fine-tune a model that can outperform the adult ASR model?
\end{enumerate}

To answer these questions, we leverage self-supervised adult speech representations to fine-tune several children’s speech recognition models on three popular native and non-native children’s corpora.

The rest of this paper is structured as follows. Section 2 describes the speech corpora that were used in our experiments. The architecture of our proposed children’s speech recognition model and methodology used to answer our research questions are described in Section 3. Experiments and results are presented in Section 4, followed by discussion in Section 5. Section 6 concludes.

\section{Speech Corpora}
\label{sec:SpCor}
Three well-known children’s speech corpora were used for fine-tuning. My Science Tutor (MyST) \cite{ward2019my} and Oregon Graduate Institute (OGI) Kids’ Speech Corpus \cite{shobaki2000ogi} comprise of speech from native English-speaking school children in America. Trentino Language Testing School (TLT-school) corpus \cite{gretter2020tlt} is a non-native English-speaking corpus which comprises of speech from Italian school children who are learning English and German. The TLT-school dataset is further split into TLT17 and TLT1618, where TLT17 includes speech recorded in 2017 and TLT1618 includes speech recorded in 2016 and 2018.

The MyST corpus contains spontaneous and prompted speech of children interacting with a virtual science tutor. The OGI corpus contains both scripted and spontaneous speech, and the TLT corpus contains prompted speech of children responding to questions from a language proficiency test.

For MyST and OGI, given the absence of official development and test sets, we crafted these sets balanced such that there is an equal distribution of speaker ages, an equal distribution of data between development and test sets for the same corpus, and speakers which appear in the development or test set do not appear in any other set. For TLT-school, we used the test set provided in the Interspeech TLT2020 challenge \cite{gretter2020overview} and used the remainder as training. Furthermore, due to memory limitations, audio samples in the training set that were greater than 15 seconds were not used. The 15 second constraint means that the training set of OGI only contains scripted child speech with limited vocabulary and variation.

The speech transcriptions were cleaned such that all transcriptions are converted to uppercase, and all punctuation except for the apostrophe are removed. Non-English words and characters that appear in the TLT-school corpus are replaced with the <unk> token to model non-English speech.

The pre-trained wav2vec2-base (BASE) model in \cite{baevski2020wav2vec} was used to extract the self-supervised speech representations. The BASE model is pre-trained on the Librispeech-960 (LS-960) corpus \cite{panayotov2015librispeech} containing speech from native English-speaking adults reading aloud.

An overview of the final dataset is shown in Table \ref{tab:table1}. For completeness, details of the adult LS-960 corpus that was used for the pre-trained wav2vec 2.0 models are also included.

\begin{table}[ht]
    \centering
    \caption {Overview of the Dataset Design}
    \includegraphics[width=0.75\textwidth]{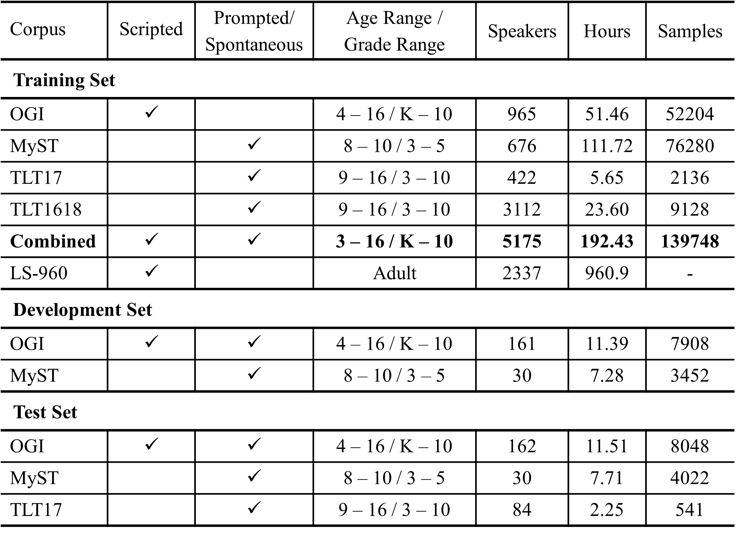}
    \label{tab:table1}
\end{table}

\section{Methodology}
\label{sec:meth}
\begin{figure}[ht]
    \centering
    \includegraphics[width=\textwidth]{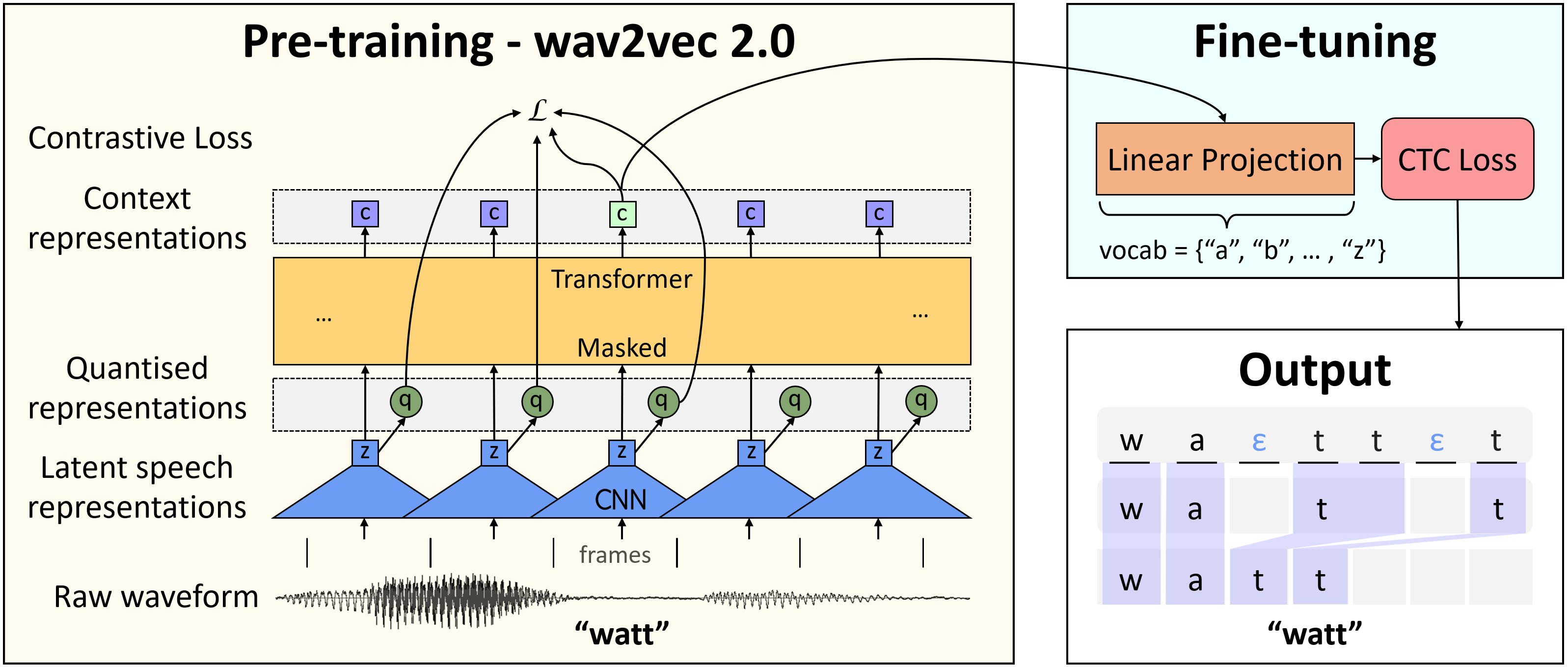}
    \caption{The proposed architecture for children’s ASR}
    \label{fig:fig1}
\end{figure}

Fig. \ref{fig:fig1} shows the architecture of our proposed children’s speech recognition system. It uses a pre-trained wav2vec 2.0 model \cite{baevski2020wav2vec} to extract contextualized speech representations. This wav2vec 2.0 model is fine-tuned for children’s speech recognition by adding a randomly initialized linear projection to predict characters, where the classes in the projection represents the vocabulary of the task. The model is optimized using the CTC algorithm \cite{graves2006connectionist}.

The output vocabulary of the linear projection consists of 26 tokens for the English alphabet characters, plus a word boundary (space) token and apostrophe token. Additionally, the <unk> token is used to represent characters that are out of the English alphabet and used to model non-English speech.

For simplicity, we did not add a language model (LM) keeping the focus of this paper on investigating the performance of self-supervised speech representations. Another reason is that the work in \cite{shivakumar2022end} which explored end-to-end children’s ASR systems found that most of the best results were obtained without an LM.

Like the work in \cite{baevski2020wav2vec}, our model is optimized with Adam \cite{kingma2014adam}, which is an adaptive learning rate optimization algorithm. The learning rate is warmed up for the first 10\% of updates, and then linearly decayed for the remainder. The learning rates and hyperparameters are chosen to align with those in \cite{baevski2020wav2vec}.

We compared the performance of our fine-tuned children’s speech recognition models to several different baselines. We used two adult baselines, wav2vec 2.0 BASE-960 and DeepSpeech 0.9.3, and evaluated them on the children’s test sets. The BASE-960 model is both pre-trained and fine-tuned on the adult LS-960 corpus and is used as the main adult baseline. DeepSpeech 0.9.3 is the most recent release of the DeepSpeech model in \cite{hannun2014deep}, which is a supervised end-to-end deep learning model built with Recurrent Neural Networks and trained with LS-960, Fisher, Switchboard, Common Voice English and transcribed radio show speech. DeepSpeech is used to compare the self-supervised adult BASE-960 model with a well-known supervised model in literature. As a child baseline, we reference results from \cite{shivakumar2022end} and \cite{knill2020non} which present the most recent work on children’s ASR: \cite{shivakumar2022end} achieved the best result for MyST (without an LM) by fine-tuning a supervised Transformer and CTC end-to-end neural network that was pretrained on Librispeech and Librivox, \cite{shivakumar2022end} achieved the best OGI performance (without an LM) by fine-tuning a supervised Time-Depth Separable and CTC end-to-end neural network that was pretrained on Librispeech and Librivox, \cite{knill2020non} on presented results for the TLT17 corpus (with an LM) using factorized time-delay neural networks. However, it is important to note that the MyST and OGI child baselines are evaluated on test sets that are designed differently compared to the ones of this paper which prevent an exact comparison. All models in this work and referenced works are evaluated using the conventional method of calculating word error rate (WER).

The experiments are performed on four Tesla V100-SXM2 32GB GPUs with a batch size of 8 per GPU, giving a total batch size of 32 samples. The CNN feature encoder component of wav2vec 2.0 is not updated during fine-tuning. Training is performed on Katana \cite{katana}, the shared computational cluster located at the University of New South Wales. All code from this paper is available at \url{https://github.com/monomest/Thesis}.

\subsection{Effect of fine-tuning}
\label{subsec:effFT}
To answer our first question of whether using child speech to fine-tune a pre-trained self-supervised model can give acceptable performance for children’s ASR, we fine-tuned the BASE model on the MyST, OGI and TLT corpora. Two native children’s models were fine-tuned using MyST and OGI corpora, which we refer to as the MyST and OGI model respectively. Two non-native children’s models were fine-tuned using the TLT corpus: one model used the small 5.65-hour TLT17 dataset (referred to as TLT17 model), and one model used both the TLT17 and TLT1618 datasets for fine-tuning (referred to as TLT model). To address the presence of non-English utterances in the datasets, these utterances were mapped to the <unk> token to model non-native speech.

The MyST model uses the same hyperparameters as the BASE model fine-tuned on 100 hours of Librispeech in \cite{baevski2020wav2vec}. The OGI model uses the same hyperparameters as the BASE model fine-tuned on 10 hours of Librispeech in \cite{baevski2020wav2vec}, but the steps increased from 20k steps to 35k steps and the learning rate decreased from 5e-5 to 4e-5, determined empirically to accommodate the larger OGI dataset. The hyperparameters for the TLT17 model were chosen to align with the model in \cite{baevski2020wav2vec} that was fine-tuned on 10 hours of Librispeech. The TLT model also used the 10-hour Librispeech hyperparameters, but the steps increased to 35k steps, and the learning rate decreased to 4e-5 to account for the larger number of hours in the corpora.

\subsection{Effect of cross-domain child corpora}
\label{subsec:effCD}
To evaluate whether ASR performance can be improved by using cross-domain child speech corpora, we fine-tuned the BASE model with several cross-domain corpora containing different combinations of child speech.

The effect of supplementing a cross-domain child corpus with adult speech is also explored by fine-tuning a model that was already fine-tuned on adult speech. Fine-tuning a fine-tuned model effectively means fine-tuning on a combined corpus of child and adult speech. To do this, we used the wav2vec2-base-960 (BASE-960) model \cite{baevski2020wav2vec} that was pre-trained and fine-tuned on the LS-960 corpus.

To explore the performance of using cross-domain adult corpora for pre-training, the wav2vec2-large-robust (ROBUST) model \cite{park2019specaugment} was used. In addition to reading-aloud speech, the ROBUST model is trained on noisy, conversational telephone data from adults. We fine-tuned the ROBUST model with cross-domain children’s corpora.

The hyperparameters for these models were chosen to align with the values for the model in \cite{baevski2020wav2vec} that is fine-tuned on 100 hours of Librispeech, with the only modification being increasing the number of training steps from 50k to 60k to accommodate for the slightly larger dataset size.

A total of five models were built, as summarized below:

\begin{enumerate}
    \item XD-native (Cross-domain native): Cross-domain model which uses the pre-trained BASE model to fine-tune on the combined MyST and OGI corpora.
    \item XD-S (Cross-domain small): Cross-domain model which uses the pre-trained BASE model to fine-tune on the combined MyST, OGI and TLT17 corpora.
    \item XD-L (Cross-domain large): Cross-domain model which uses the pre-trained BASE model to fine-tune on the larger cross-domain corpus combining MyST, OGI, TLT17 and TLT1618 corpora.
    \item XDL-adult (Cross-domain large-adult): Cross-domain model supplemented by adult speech, which uses the BASE-960 model to fine-tune on combined MyST, OGI, TLT17 and TLT1618 corpora.
    \item XDL-robust (Cross-domain large-robust): Cross-domain model which uses the pre-trained ROBUST model to fine-tune on combined MyST, OGI, TLT17 and TLT1618 corpora.
\end{enumerate}

\subsection{Effect of data quantity}
\label{subsec:data}
The effect of the amount of data available for fine-tuning on ASR performance was investigated by fine-tuning a children’s ASR model with various amounts of MyST data. Subsets of 10 minutes, 1 hour, 5 hours, 10 hours and 110 hours were used.

The relationship between data availability and ASR performance across children’s ages was also assessed. We split the OGI test set based on the grades of the speakers, and the WER of the XD-L, XDL-adult, and OGI models were obtained for each grade. An additional model “XD-L half” was fine-tuned with half the number of hours of the XD-L model to further illuminate the effect of data quantity. The OGI model was evaluated because its fine-tuning data consists of an approximately equal spread of ages across all grades, whereas the XD models include the MyST and TLT-school datasets which have a limited age range. The OGI corpus was used because this is the only corpus which contains grade-level information in the speech files.

Model performance for speech samples with different utterance lengths (or number of words in a speech sample), and its relationship with data quantity was investigated. We did this by grouping the MyST test set into subgroups based on utterance lengths, and then evaluating the performance of the XD-L and XDL-adult models for each subgroup. The MyST corpus was used because it contained the most variations in utterance lengths.

As an additional experiment, the effectiveness of data augmentation using SpecAugment \cite{park2019specaugment} was verified. A cross-domain children’s model was fine-tuned without applying the SpecAugment technique, referred to as XDL-noSpec (cross-domain with no SpecAugment). XDL-noSpec is the same cross-domain model as XD-L but trained without applying SpecAugment.  Besides this experiment, all other models applied SpecAugment.

\section{Results}
\label{sec:res}

\subsection{Effect of fine-tuning}
\label{subsec:effFT_res}

\begin{figure}[ht]
    \centering
    \includegraphics[width=0.75\textwidth]{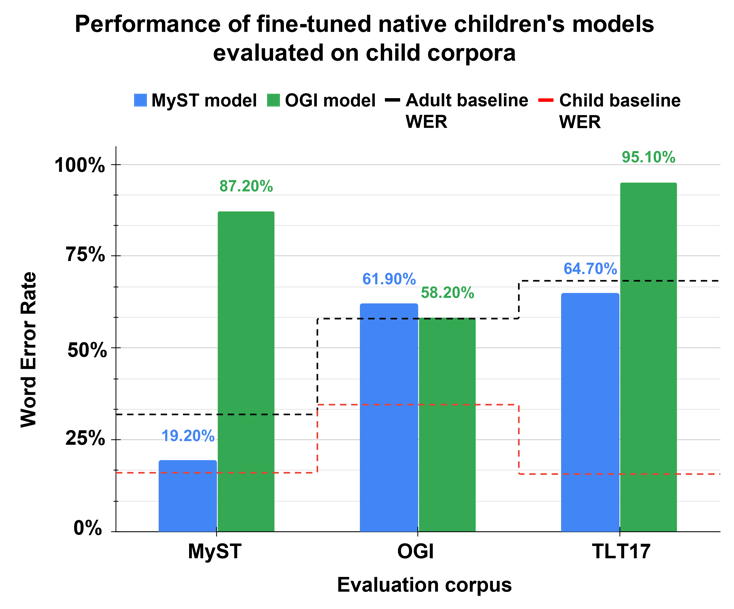}
    \caption{Performance of the fine-tuned native child models, evaluated on the MyST, OGI and TLT17 test set. The fine-tuned MyST model outperforms the adult baseline for both MyST and TLT17 test sets, however no model performs better than the child baselines of each test set.}
    \label{fig:fig2}
\end{figure}

\begin{figure}[ht]
    \centering
    \includegraphics[width=0.75\textwidth]{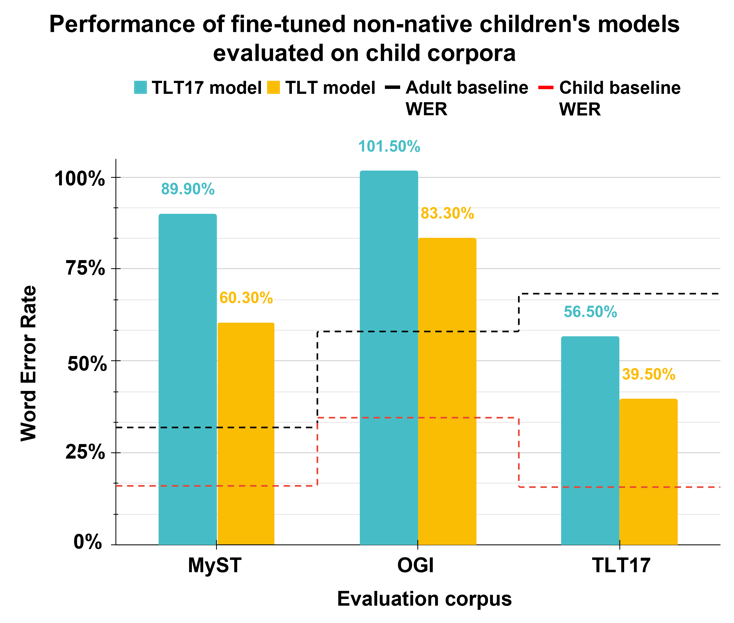}
    \caption{Performance of the fine-tuned non-native child models, evaluated on the MyST, OGI and TLT17 test set. Expectedly, both non-native models perform poorly on the native MyST and OGI test sets, however the fine-tuned TLT model significantly outperforms the adult BASE-960 baseline for the non-native TLT17 test set.}
    \label{fig:fig3}
\end{figure}

Figs. \ref{fig:fig2} and \ref{fig:fig3} summarize the performance on our fine-tuned models evaluated on different child corpora, and Table \ref{tab:table2} presents the full results. The adult baseline WERs are obtained by evaluating the baseline adult model BASE-960 on the various test sets. The MyST and OGI child baseline WERs are referenced from the best results in \cite{shivakumar2022end} for models which do not use an LM. The TLT17 child baseline WER is referenced from the best results in \cite{knill2020non}, which does use an LM.

\begin{table}[ht]
    \centering
    \caption{Performance of Fine-tuned Children’s Models}
    \includegraphics[width=0.75\textwidth]{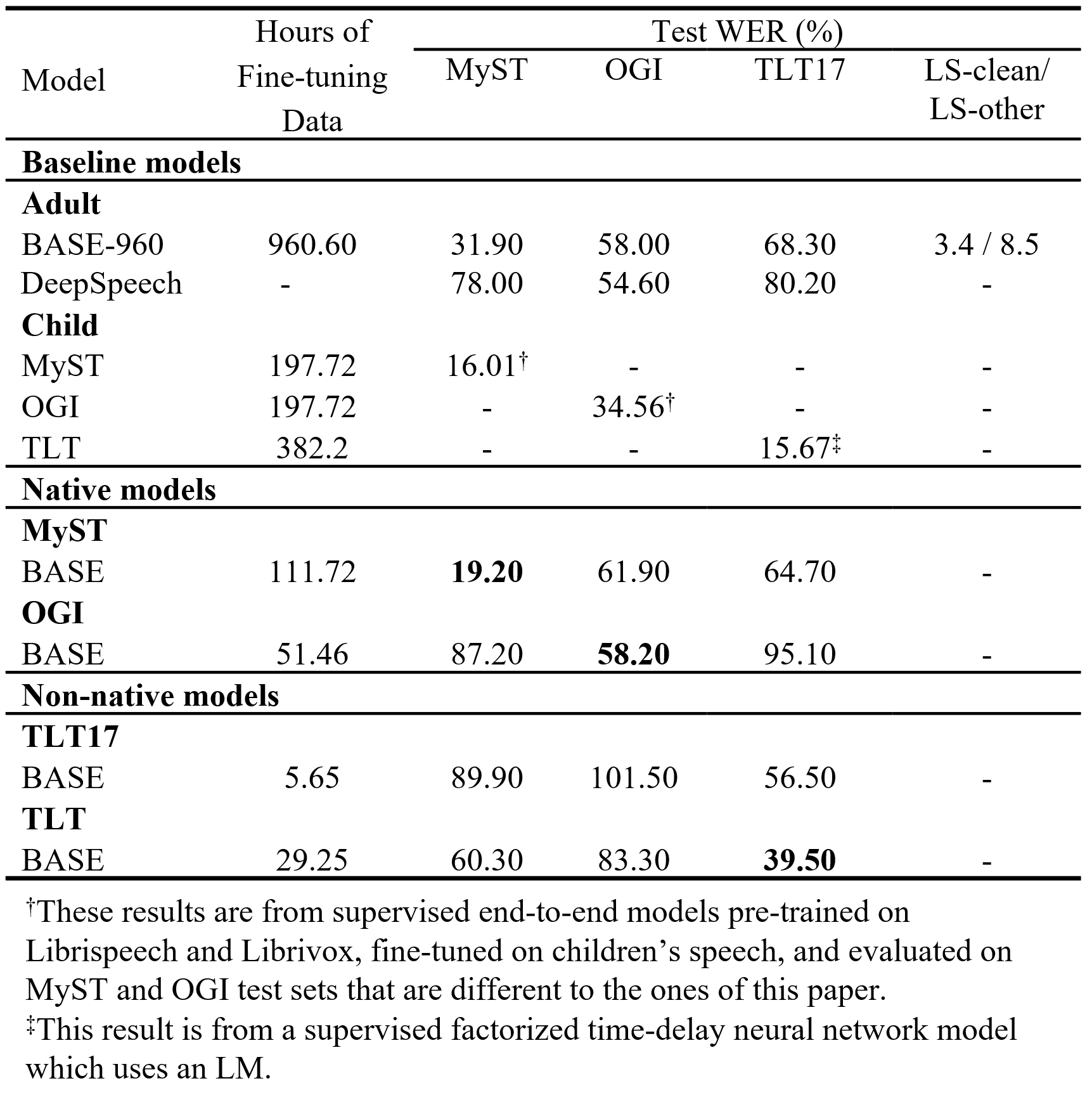}
    \label{tab:table2}
\end{table}

First, we observed that the best results for each test corpus come from models which are fine-tuned on the same corpus, and that these results often outperform the adult baselines but do not outperform the supervised child baseline models. As seen in Table \ref{tab:table2}, the best performing model for the MyST test set is the MyST model with a WER of 19.20\%. This result yields a 12.70\% absolute improvement and 39.81\% relative improvement from the adult BASE-960 baseline WER. However, the MyST model underperforms the child baseline in \cite{shivakumar2022end} by 3.19\% absolute and 19.92\% relative WER. Similarly, the best model for the OGI corpus is the OGI model with a WER of 58.20\%. This is result is slightly worse than the adult BASE-960 baseline by 0.20\%. This is explained by the analysis in Section 4.3.1. The OGI model overfitted the scripted speech component of the OGI corpus and thus the OGI model achieved very low WER for this part, However, the limited vocabulary and acoustic variation in the OGI fine-tuning set led to very high WER for the spontaneous speech component of the test set.  Fig. \ref{fig:fig3} shows that the best model for the non-native TLT17 corpus is the TLT model (39.50\% WER) which is fine-tuned on all the available non-native speech. Increasing the fine-tuning data from 5.65 hours of TLT17 to 29.25 hours of the TLT model yields an absolute WER improvement of 17\% and a 30\% relative improvement. Although the TLT model has a 28.80\% absolute WER improvement compared to the adult BASE-960 baseline, it still underperforms the child baseline by 23.83\% absolute WER.

\subsection{Effect of cross-domain child corpora}
\label{subsec:effCD_res}
\begin{figure}[ht]
    \centering
    \includegraphics[width=0.75\textwidth]{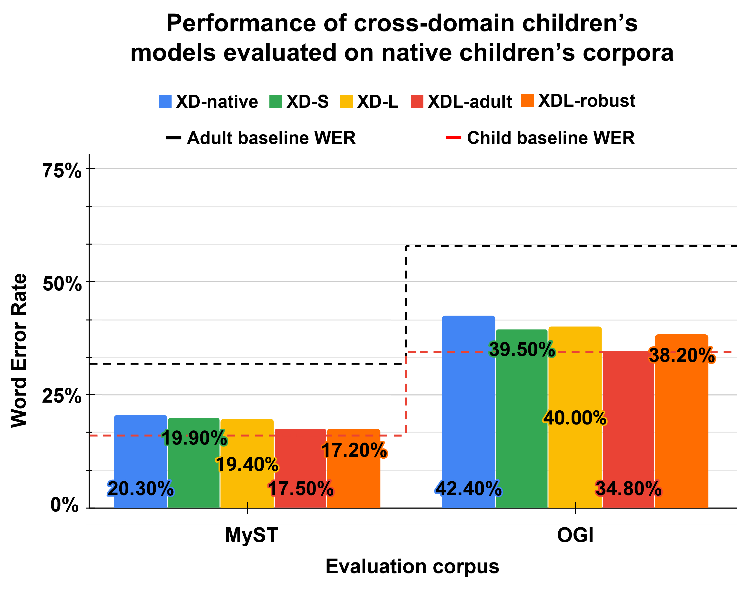}
    \caption{Five cross-domain child models (XD-native, XD-S, XD-L, XDL-adult, XDL-robust) were evaluated on the native MyST and OGI test sets. All models perform significantly better than the adult BASE-960 baselines and are comparable to child baselines for both native test sets.}
    \label{fig:fig4}
\end{figure}

\begin{figure}[ht]
    \centering
    \includegraphics[width=0.75\textwidth]{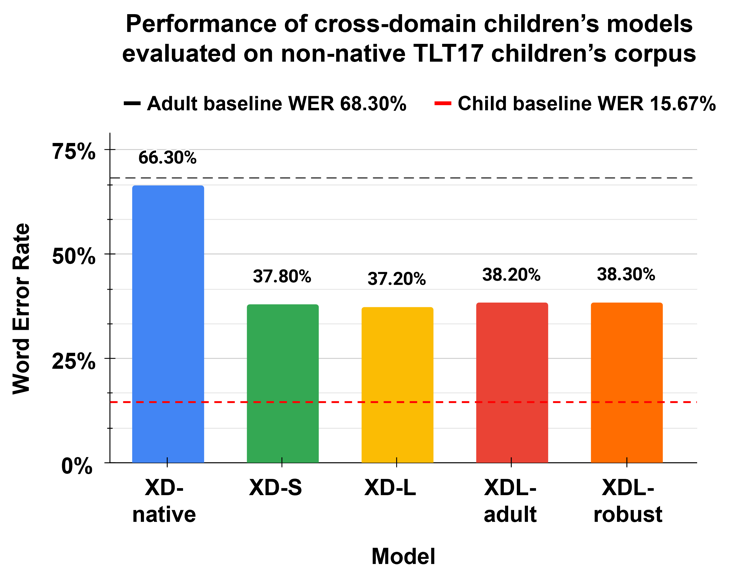}
    \caption{Five cross-domain child models (XD-native, XD-S, XD-L, XDL-adult, XDL-robust) were evaluated on the non-native TLT17 test set. The best performing model for non-native children’s ASR across all single-domain and cross-domain models is XD-L, which was trained with both native and non-native child speech.}
    \label{fig:fig5}
\end{figure}

The results of our cross-domain models are summarized in Figs. \ref{fig:fig4} and \ref{fig:fig5}, and the full results are shown in Table \ref{tab:table3}. The adult and child baselines are the same as in Section 4.1.

\begin{table}[h!]
    \centering
    \caption{Results of Cross-Domain Children’s Models}
    \includegraphics[width=0.75\textwidth]{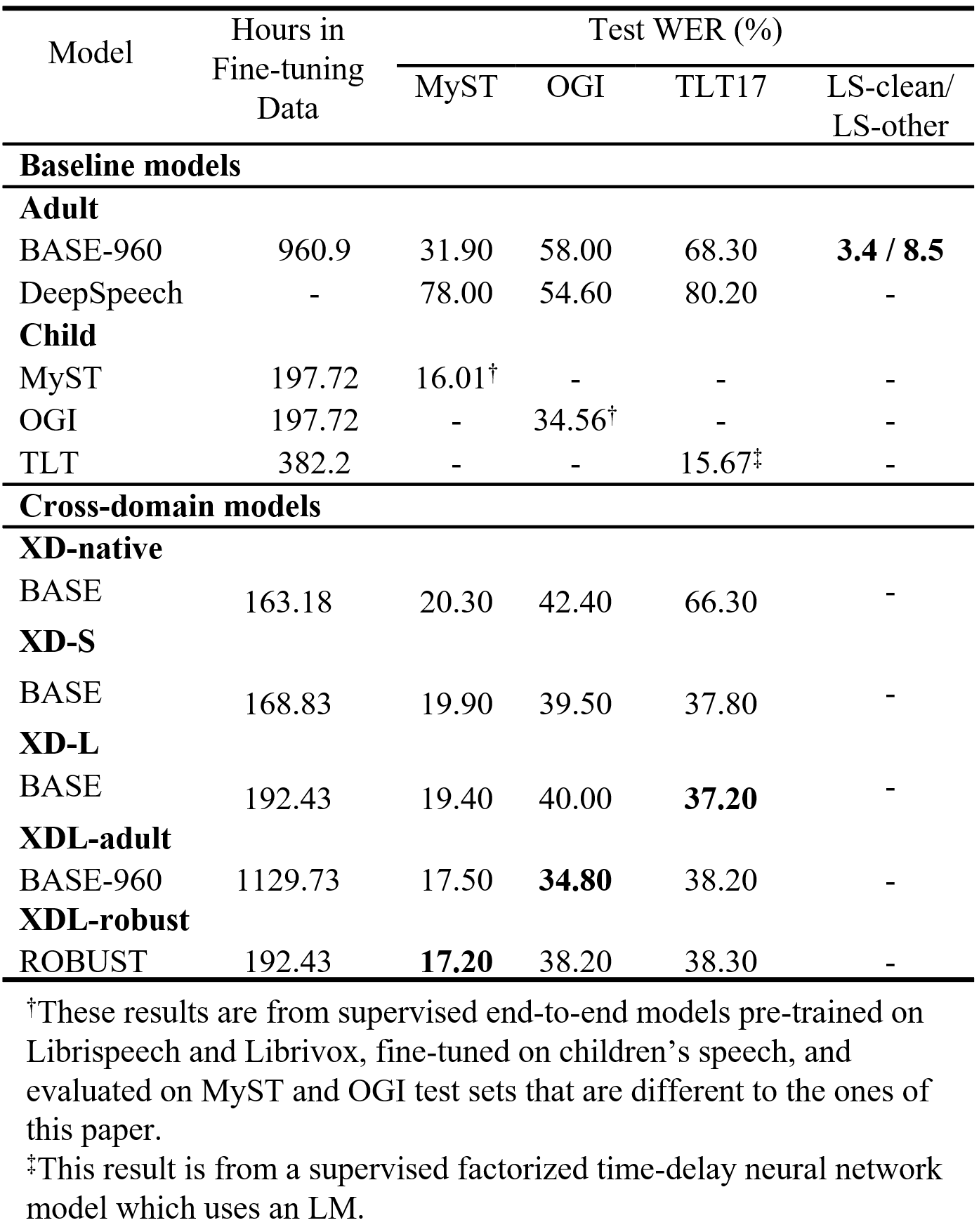}
    \label{tab:table3}
\end{table}

All our models considerably outperform the baseline adult models for all test sets and native models are comparable to the supervised child baselines. In Fig. \ref{fig:fig4}, the best result for the MyST test set is from the XDL-robust model with 17.20\% WER, 14.70\% absolute improvement and 46.08\% relative improvement compared to the adult BASE-960 baseline. This is comparable to the child baseline WER of 16.01\% from \cite{shivakumar2022end}. Furthermore, Fig. \ref{fig:fig4} shows that the XDL-adult model gave the best OGI test results with 34.80\% WER, 23.20\% absolute improvement and 40\% relative improvement compared to the adult BASE-960 baseline. Again, this result is comparable to the 34.56\% WER of the child baseline in \cite{shivakumar2022end}.

In Fig. \ref{fig:fig5}, the best result for the non-native TLT17 test set was from the XD-L model with 37.20\% WER, 31.10\% absolute improvement and 45.50\% relative improvement compared to the adult BASE-960 baseline. The XD-L model underperforms the child baseline in \cite{knill2020non} by 21.53\% absolute WER, possibly because the work in \cite{knill2020non}  included a language model, handled learner pronunciation errors, and used grade specific models to account for age variations.

Cross-domain models also outperformed the single-domain models in Section 4.1. Comparing the best cross-domain models to the best single-domain models, cross-domain models gave an absolute and relative WER improvement of 2\% and 10.41\% for the MyST corpus, 23.40\% and 40.20\% for the OGI corpus, and 2.30\% and 5.82\% for the TLT17 corpus.

Furthermore, Fig. \ref{fig:fig4} reveals that the XD-L model which is fine-tuned on both native and non-native child speech outperforms the native-only XD-native model for the task of native children’s ASR. The XD-L model gives an absolute WER improvement of 0.90\% and 2.40\% for the MyST and OGI corpus respectively, compared to the XD-native model.

As shown in the Fig. \ref{fig:fig4}, the inclusion of adult speech in XD-adult leads to a 9.8\% and 13\% relative performance improvement for the MyST and OGI corpora respectively, in comparison to XD-L. However, Fig. \ref{fig:fig5} shows that for non-native children’s ASR for TLT17, the inclusion of adult speech for fine-tuning gives a 1\% decrease in performance compared to XD-L.

In Table \ref{tab:table3}, we see that the ROBUST model leads to some improvement for the MyST corpus compared to the best performing BASE model, however this is minor (0.30\% absolute and 1.7\% relative WER reduction). In contrast, the ROBUST model underperforms for the OGI and TLT17 corpus. A comparison between the adult BASE-960 baseline model’s performance for adult speech (3.40\% and 8.50\% WER) and the best performance for child speech (17.50\% WER for MyST) reveals that children’s ASR is still not on par with the adult ASR.

\subsection{Effect of data quantity}
\label{subsec:data_res}

\begin{figure}[ht]
    \centering
    \includegraphics[width=0.75\textwidth]{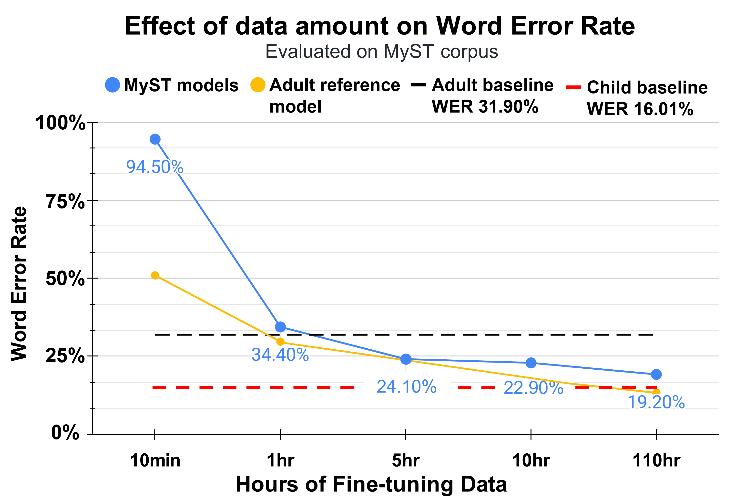}
    \caption{Performance of MyST models fine-tuned with various amounts (10 minutes, 1 hour, 5 hours, 10 hours, and 110 hours) of child speech, and evaluated on the MyST test set. The adult reference model (yellow) is from the results in \cite{baevski2020wav2vec}, the adult baseline WER (black) is the WER for MyST using the baseline adult BASE-960 model, and the child baseline WER (red) is the MyST result from \cite{shivakumar2022end}. Using only 110 hours of fine-tuning data, it is possible to get comparable results to the child baseline model that is trained with 192 hours.}
    \label{fig:fig6}
\end{figure}

Fig. \ref{fig:fig6} presents the results of the MyST models that were fine-tuned on different sized subsets of the MyST corpus and evaluated on the MyST test set. As an adult reference model, the results from \cite{baevski2020wav2vec} which performed a similar experiment using adult speech are also plotted. In a low resource set-up with only 1 hour of transcribed children’s speech the MyST model reaches 34.40\% WER, which is near the adult BASE-960 baseline performance of 31.90\%. Once the fine-tuning data increases to 5 hours, the MyST children’s model surpasses the adult BASE-960 baseline by 7.80\%. As the amount of fine-tuning data increases, the MyST model performance reaches closer to the child baseline of 16.01\% WER in \cite{shivakumar2022end} which was obtained using 197.72 hours child speech.

\subsubsection{Age performance}
\label{subsubsec:age_res}
The performance of XD-L, XDL-adult, XD-L half, OGI and the adult BASE-960 baseline model across different school grades are shown in Table 4. Table 5 separately evaluates the models’ performance for scripted and spontaneous speech in the OGI test set.

\begin{table}[ht]
    \centering
    \caption{Overall Performance Across Age for OGI dataset}
    \includegraphics[width=0.75\textwidth]{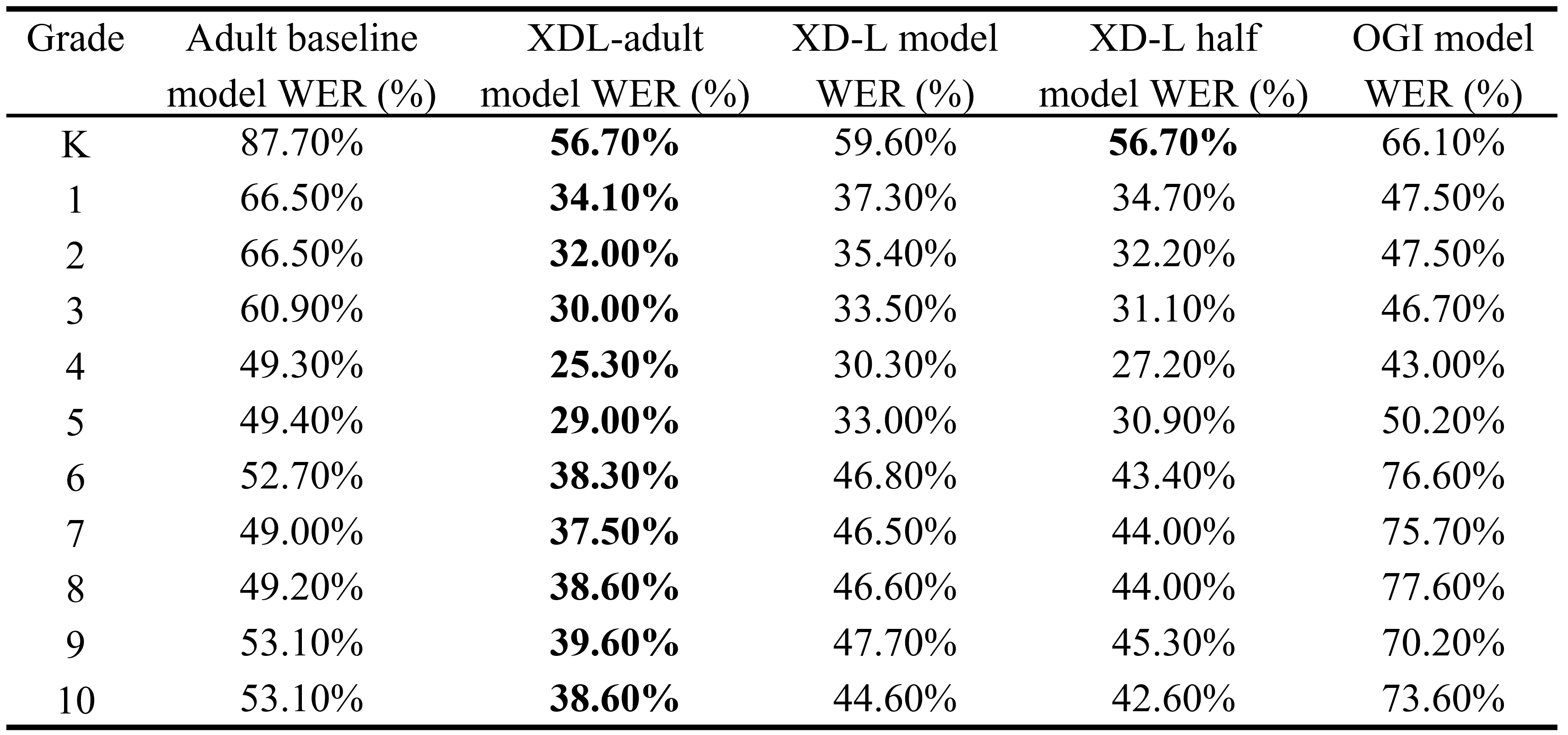}
    \label{tab:table4}
\end{table}

\begin{table}[ht]
    \centering
    \caption{Performance Across Age separated by Scripted and Spontaneous Speech for OGI dataset}
    \includegraphics[width=0.75\textwidth]{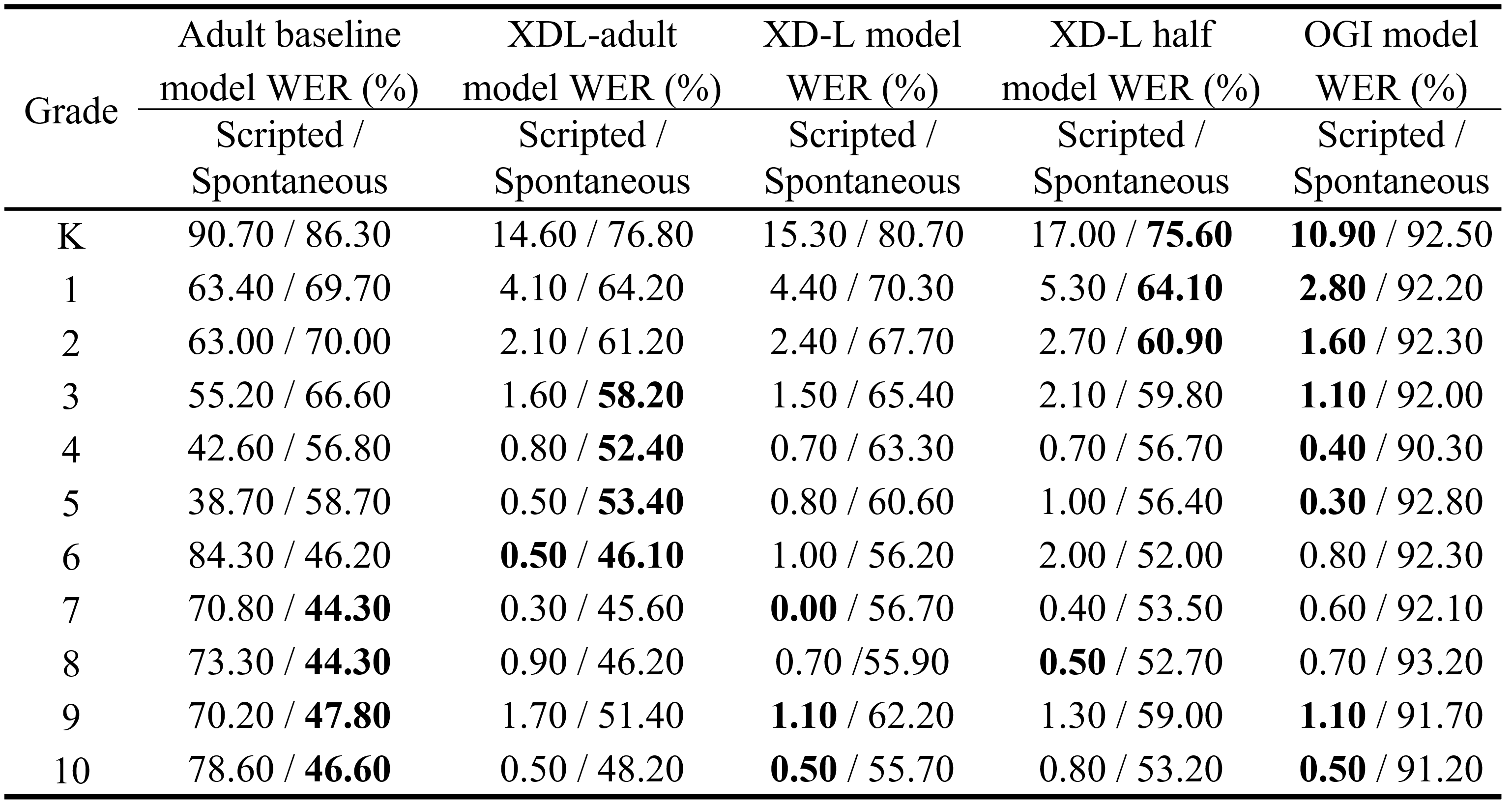}
    \label{tab:table5}
\end{table}

Tables \ref{tab:table4} and \ref{tab:table5} show that the most challenging age range is Kindergarten students (56.70\% best WER overall) however there is a wide performance gap between scripted and spontaneous speech (10.90\% best WER scripted, 75.60\% best WER spontaneous). For all models, there is a steep drop in WER from Kindergarten to Grade 1 and a general downward trend in WER as age increases. The best overall age was Grade 4 (25.30\% WER), while scripted speech had the best performance in Grade 7 (0.00\% WER) and spontaneous in Grade 7 and 8 (44.30\%).

Interestingly, comparing the XD-L and XD-L half models shows that doubling the amount of children’s fine-tuning data only improves age performance for scripted speech, and worsens performance for spontaneous speech.

Although the XDL-adult model achieved the best overall results for every age, the OGI model is superior for Grades K to 5 in recognizing scripted speech. The adult BASE-960 baseline model fine-tuned without children’s speech consistently outperforms all other children’s models for spontaneous speech in Grades 7 to 10, however it is inferior in all other settings.

\subsubsection{Utterance Length}
\label{subsubsec:uttL_res}
\begin{figure}[ht]
    \centering
    \includegraphics[width=0.75\textwidth]{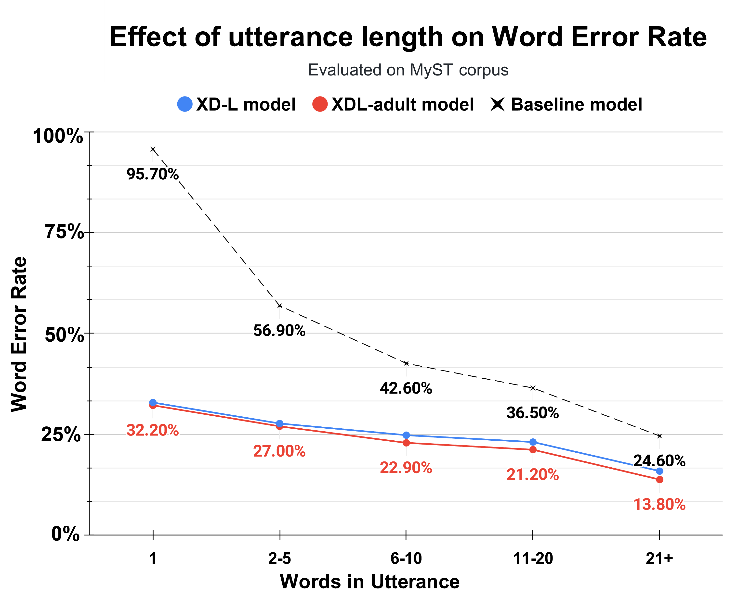}
    \caption{Performance of XD-L, XDL-adult and adult BASE-960 baseline models evaluated on the MyST test set, grouped by utterance length. Performance improves as utterance length increases and resembles the long sentences contained in the Librispeech pre-training corpus.}
    \label{fig:fig7}
\end{figure}

The results of evaluating XD-L, XDL-adult and the baseline adult model for various utterance lengths in the MyST corpus are presented in Fig. \ref{fig:fig7} and Table \ref{tab:table6}. All models perform better for longer utterance lengths, while shorter utterance lengths, especially those with only one word, are more challenging.

\begin{table}[ht]
    \centering
    \caption{Performance Across Utterance Lengths}
    \includegraphics[width=0.75\textwidth]{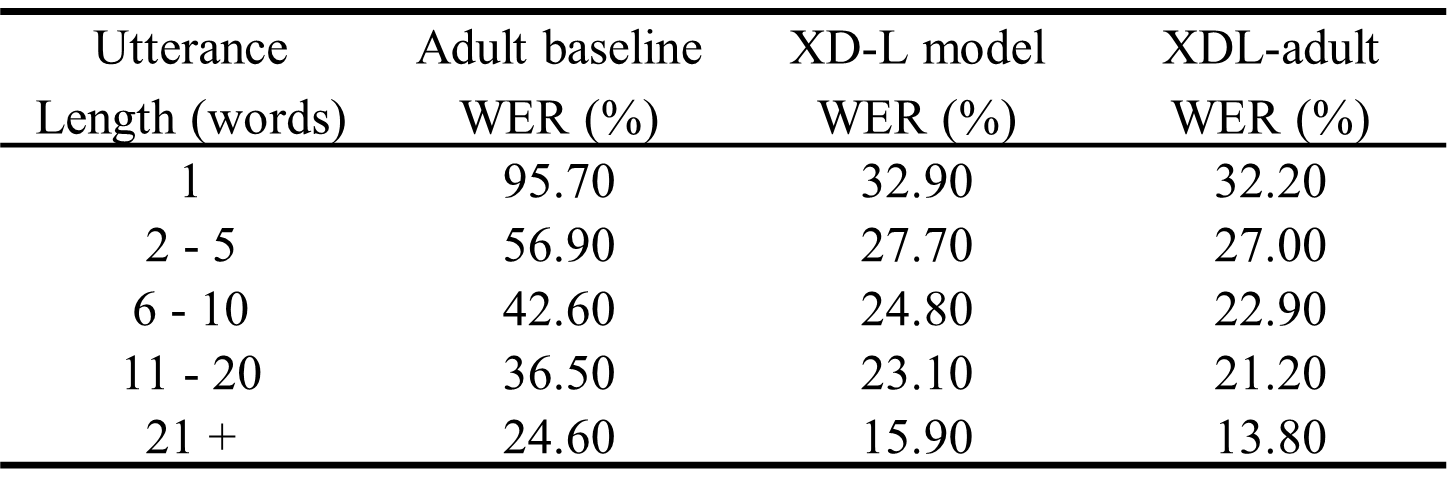}
    \label{tab:table6}
\end{table}

\subsubsection{SpecAugment}
\label{subsubsec:spec_res}

Fig. 8 shows the results of models with and without SpecAugment. Fig. 8 shows that applying SpecAugment consistently improves results. SpecAugment gives a minimum absolute improvement of 0.60\% and minimum relative improvement of 3\% when evaluated on the MyST corpus and leads to a maximum absolute improvement of 3.60\% and maximum relative improvement of 8.90\% when evaluated on the TLT17 corpus.

\begin{figure}[ht]
    \centering
    \includegraphics[width=0.75\textwidth]{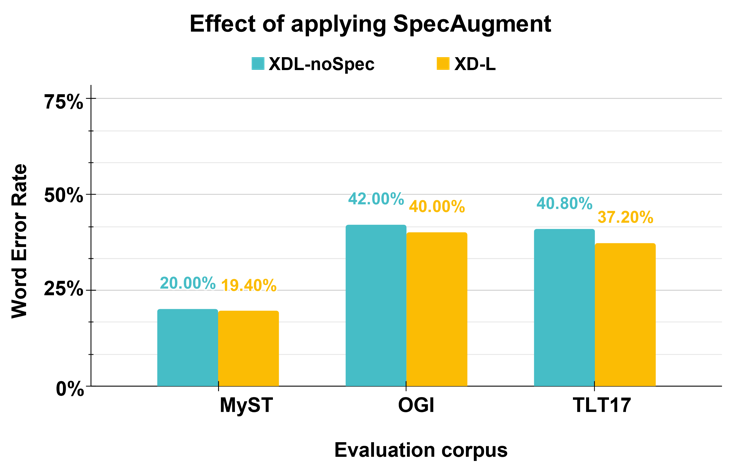}
    \caption{Performance of cross-domain XDL-noSpec model which does not apply SpecAugment, and cross-domain XD-L model which does apply SpecAugment. Applying SpecAugment improves performance for all test sets.}
    \label{fig:fig8}
\end{figure}

\section{Discussion}
\label{sec:dis}
\subsection{Effect of fine-tuning}
\label{subsec:effFT_dis}
As seen in Table \ref{tab:table2}, fine-tuning on the MyST corpus yields significant improvements for the MyST corpus compared to the baseline adult models, however this level of improvement is not experienced by the other children’s datasets. Surprisingly, the MyST corpus also generated minor improvements for the non-native TLT17 corpus. However, the fine-tuned MyST model underperformed the adult baselines when evaluated on the OGI corpus. This suggests that although single-domain models can do extremely well for the dataset of interest, performance improvements using fine-tuned self-supervised models are not generalizable to other domains and datasets. A similar observation has been made in \cite{shivakumar2022end} for supervised transformer-based end-to-end neural ASR systems for children.

Table \ref{tab:table2} also reveals that the fine-tuned OGI model underperforms all baselines when evaluated on every test corpus, including OGI itself. However, when assessing the OGI model’s performance on the OGI scripted test set in isolation (Table \ref{tab:table5}), OGI generally outperforms all other models. A likely explanation is that the OGI fine-tuning set has limited vocabulary and acoustic variation, which leads to overfitting on scripted speech. Additional reasons could be the long read-aloud utterances and adult age range in the Librispeech pre-training corpus, which is dissimilar to the OGI child corpus. This implies that obtaining data from a diversity of speakers should be done in the pre-training stage. If the fine-tuning data has a similar data distribution to the real application, self-supervised models are very likely to assist the final performance of an ASR system.

Furthermore, it is promising that the MyST model is somewhat comparable to a similar adult LS model from \cite{baevski2020wav2vec}. The MyST model fine-tuned on 110 hours of speech gave a MyST WER of 19.20\%, and an adult model fine-tuned on 100 hours of Librispeech gave a 13.30\% WER on the LS-other test set. While the adult model is pre-trained, fine-tuned and evaluated using the same domain of adult data, the MyST children’s model uses a different domain and corpus for fine-tuning and evaluation. Furthermore, the longer utterances in read-aloud adult speech, and age-related variations which pervade child speech but are not present in adult speech also cloud naïve comparisons between child and adult model performances. Nonetheless, these comparable results show that using child speech to fine-tune on pre-trained self-supervised adult models can be effectively used for children’s ASR despite the domain mismatch.

Additionally, the self-supervised BASE-960 adult baseline far surpassed the supervised DeepSpeech adult baseline for children’s ASR performance. Although BASE-960 was only trained on one adult corpus (LS-960) it is more robust to children’s speech than DeepSpeech, which is trained on 5 adult corpora including LS-960. This suggests even without fine-tuning, using a pre-trained self-supervised model could be more advantageous than a pre-trained supervised model when performing ASR for out-of-domain speech.

Unfortunately, the non-native children’s ASR is not comparable to the native adult ASR. This implies that the adaptability of the fine-tuned model to non-native speakers is limited by the performance of the self-supervised speech representations.

\subsection{Cross-domain children’s models}
\label{subsec:effCD_dis}
The best performance for native children’s ASR is obtained when fine-tuning on a cross-domain corpus containing all available datasets. That is, fine-tuning with both non-native child speech and native adult speech improves the performance for native children’s ASR. Such a result is promising because it means that models which use self-supervised speech representations can leverage what it learns from adult speech in both the pre-training and fine-tuning stage and apply this to enhancing the performance for children’s ASR.

Additionally, fine-tuning using a cross-domain corpus which includes both native and non-native child speech leads to the best performance for non-native children’s ASR. It is more effective to combine the 5.65 hours of non-native TLT17 with the native MyST and OGI corpora (37.80\% WER) than to use only TLT17 and TLT1618 (39.50\% WER from Table \ref{tab:table3}). This means that to build ASR models for low-resource non-native children’s speech, it is possible to supplement the limited non-native speech corpus with more abundant native children’s speech to obtain performance improvements. This suggests that architectures which utilize a self-supervised model show promise in being generalizable to other domains of speech. Our results align with those in \cite{shibano2021speech} and \cite{deng2021improving}, which focused on leveraging such architectures for non-native adult speech. These results reveal that a diverse dataset matters in improving performance. However, it is important to note that non-native children’s speech recognition is not improved by including native adult speech, possibly due to the too strongly dissimilar domains.

A comparison of the self-supervised cross-domain models with the supervised child baseline models in Table \ref{tab:table3} reveal that for native-speaker children, self-supervised models are on par with supervised pre-trained models: XDL-robust achieved 17.20\% WER in comparison to the 16.01\% WER baseline for MyST, and XDL-adult achieved 34.80\% WER in comparison to the 34.56\% baseline for OGI. Although a direct comparison cannot be made due to differences in test sets, these comparable results are a promising indicator for the efficacy of leveraging self-supervised models when tackling low-resourced domains such as children’s speech.

Unfortunately, for non-native children the best cross-domain model XD-L still severely underperforms the supervised TLT baseline model of 15.67\% WER by a factor of 1.5. The likely cause is that the baseline included an LM and accounted for both learner pronunciation errors and age variations. This means that although self-supervised models are incredibly adaptable to fine-tuning data that is of a different distribution to the pre-training data, fine-tuning self-supervised models catered to native speakers is not enough to achieve high performance for non-native children. 

Whether the ROBUST cross-domain adult wav2vec 2.0 model is superior to the single-domain BASE model is uncertain. Only the MyST corpus experienced improvements, while the ROBUST model underperforms the BASE model for the OGI and TLT17 corpus. A possible explanation is that the MyST corpus more closely resembles to the datasets used to pre-train the ROBUST model compared to the other children’s corpora.

\subsection{Effect of data quantity}
\label{subsec:data_dis}
Our results show that a minimum of 5 hours is needed to fine-tune a children’s ASR model that outperforms the adult BASE-960 baseline fine-tuned on 960 hours of adult speech. Such a result is extremely promising because it reveals that a decent ASR model can be built in low resource scenarios, by leveraging self-supervised speech representations that are pre-trained on a different domain. On the other hand, the diminishing returns to increasing the amount of fine-tuning data reveal that improvements to the performance of children’s ASR are constrained by the robustness of the self-supervised speech representations.

\subsubsection{Age performance}
\label{subsec:age_dis}
A key observation is that increasing the amount of child speech in fine-tuning only improves age performance under specific settings and has diminishing returns to scale, as shown in Fig. 9. A data increase improves performance for scripted speech across all ages yet worsens performance for spontaneous speech across all ages. A possible cause is the fact that the distribution of the fine-tuning dataset only matches the distribution of the scripted test set.

Kindergarten children are simultaneously the most challenging speakers and one of the least sensitive to data increases. Doubling the fine-tuning data gives a 10\% relative WER improvement for Kindergarten scripted speech. This suggests that achieving a similar performance between Kindergarten children (10.90\% WER in Table 5) and older children (0.50\% WER in Table 5) will require enormous amounts of fine-tuning data. Therefore, including un-transcribed speech from young children in the unsupervised pre-training stage will likely yield the most scalable and effective solution for achieving high ASR performance across all children’s ages and types of speech.

\begin{figure}[ht]
    \centering
    \includegraphics[width=0.75\textwidth]{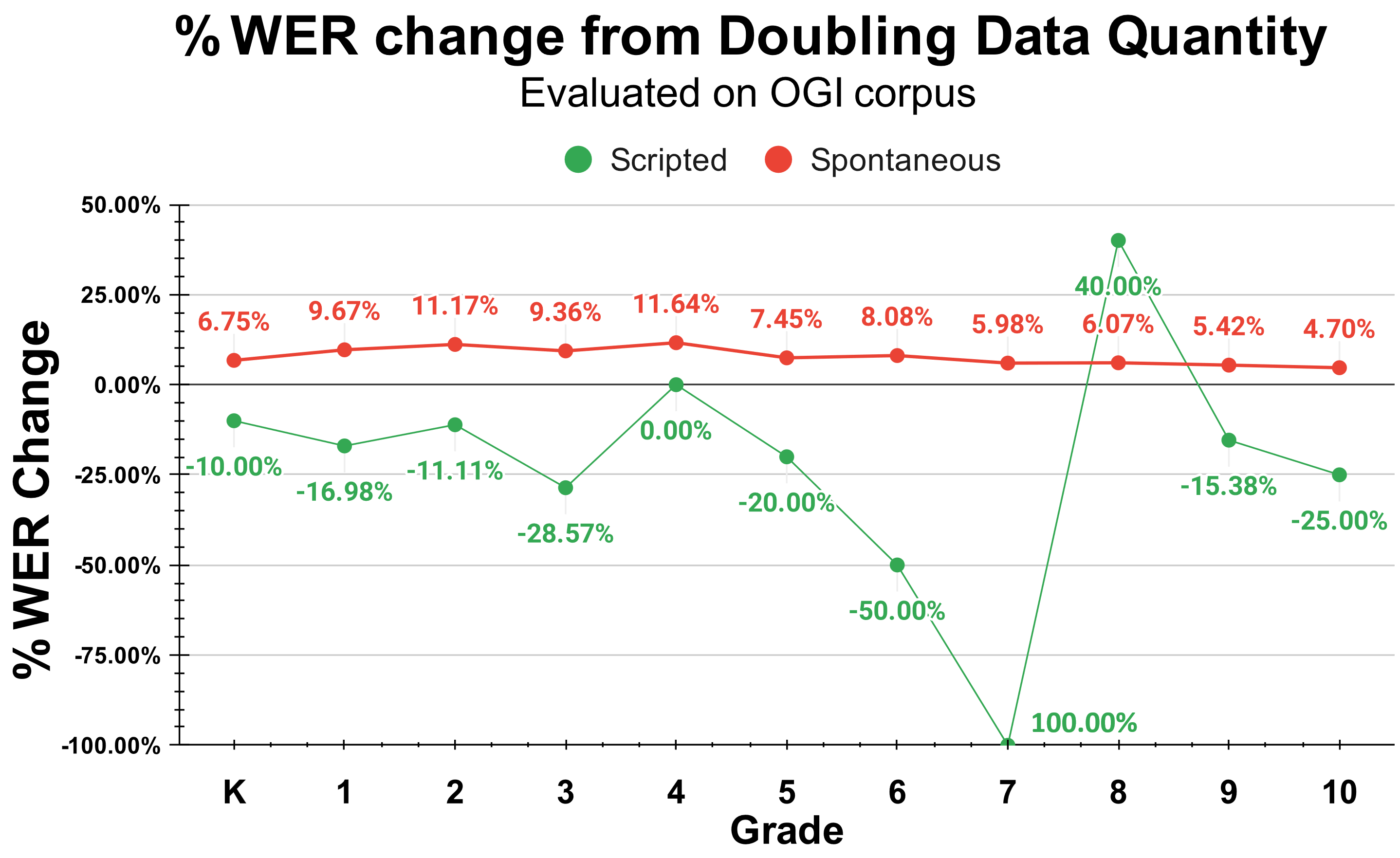}
    \caption{Percentage change in WER from XD-L half to XD-L, across children’s ages. Doubling the fine-tuning data generally decreases (improves) WER for scripted speech, although its effect varies with age. However, there is a consistent degradation in performance for spontaneous speech across all ages when fine-tuning data is doubled.}
    \label{fig:fig9}
\end{figure}

\subsubsection{Utterance Length}
\label{subsubsec:uttL_dis}
Our results show that the most challenging utterance length is one word, and performance improves as the utterance length increases. The correlation values in Table 7 reveal that the strongest correlation is between the performance of the adult BASE-960 baseline model and the fine-tuned models, rather than between model performance and data quantity. The XD-L WER has a correlation of 0.93 with the baseline adult WER. This implies that the self-supervised speech representations have a stronger influence on how fine-tuned models perform across utterance lengths, compared to data quantity for each utterance length.

The significant decrease in WER for one-word speech samples for the cross-domain models (from 95.70\% WER in the baseline to 32.20\% WER for XDL-adult) can be attributed to the large amount of data available for one-word utterance lengths, however Fig. 7 shows that the overall trend of the fine-tuned models aligns with the baseline model. The better performance for longer utterance lengths is reasonable because the self-supervised representations are pretrained using the Librispeech corpus, which contains longer utterance lengths due to the corpus being from adults reading books aloud. This makes comparing child and adult state-of-the-art ASR more negatively biased towards child ASR, as the adult Librispeech test set would more closely resemble the longer read-aloud utterances than the children’s test corpora.

\begin{table}[ht]
    \centering
    \caption{Correlation between Utterance Length and Data}
    \includegraphics[width=0.75\textwidth]{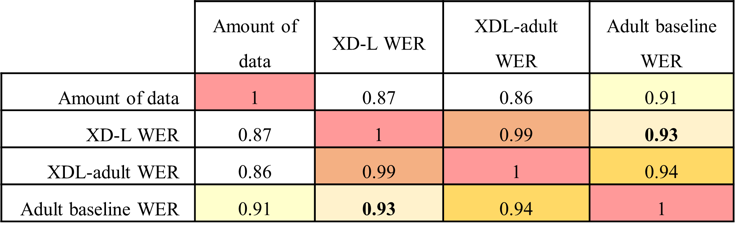}
    \label{tab:table7}
\end{table}

\subsubsection{SpecAugment}
\label{subsubsec:spec_dis}

Lastly, we verified that applying SpecAugment for data augmentation does improve ASR performance for children’s speech. As shown in Fig. 8, when SpecAugment is used with cross-domain children’s data, it yields a 2\% absolute WER improvement, and 4.76\% relative WER improvement compared to the model where SpecAugment is not applied.

\subsection{Summary of Results}
\label{subsec:sum}
The best performing models for each children’s speech corpora are summarized in Table 8.

\begin{table}[ht]
    \centering
    \caption{Summary of Best Performing Models}
    \includegraphics[width=0.75\textwidth]{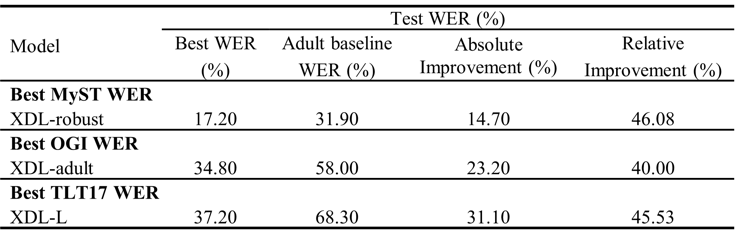}
    \label{tab:table8}
\end{table}

The speech corpus with the highest relative improvement compared to the baseline model is the MyST corpus (46.08\% improvement), using the self-supervised ROBUST speech representations model and fine-tuned on the cross-domain child corpus containing all the available children’s speech corpora. The non-native TLT17 children’s corpus obtained the highest absolute WER improvement (31.10\%), using the BASE speech representations model and fine-tuned on all available children’s speech corpora. All models experienced at least a 40\% relative improvement over the baseline adult model that is fine-tuned with 960 hours of adult speech, with less than 200 hours of child speech for fine-tuning. These significant improvements highlight the promise of self-supervised speech representations for overcoming the challenges of children’s speech.

\section{Conclusion}
\label{sec:con}
In this paper, we explored the use of self-supervised adult speech representations in building automatic speech recognition systems for children. Our proposed children’s speech recognition system used a pre-trained wav2vec 2.0 model followed by a randomly initialized linear projection that is optimized using CTC.  We fine-tuned models using native and non-native child speech from three well-known child corpora. Our experiments answered 3 key questions:

\begin{enumerate}
    \item What is the effect of fine-tuning: Fine-tuning with a single children’s speech corpus is effective when the fine-tuning data has a similar data distribution to the application and leads to performance improvements of up to 39.81\% relative improvement compared to the adult baseline for native speech, and 28.80\% for non-native speech.
    \item What is the effect of cross-domain corpora: Cross-domain fine-tuning leads to the best performance for both native and non-native child speech. For native speech, cross-domain models are comparable to other children’s models in literature and yields up to 46.08\% relative improvement from the adult baseline. For non-native speech, cross-domain models underperform those in literature but achieves 45.50\% relative improvement from the adult baseline.
    \item What is the effect of data quantity: A minimum of 5 hours is needed to fine-tune a native children’s ASR model that outperforms the adult baseline. Increasing fine-tuning data quantity improves performance across ages under certain circumstances and has diminishing returns: doubling fine-tuning data gives a 10\% relative WER improvement for Kindergarten scripted speech. Self-supervised speech representations have a stronger influence on performance across utterance lengths, compared to the quantity of fine-tuning data for each utterance length.
\end{enumerate}

These results show that self-supervised adult speech representations can be successfully exploited for children’s speech recognition and outperform high-performance adult models with minimal data requirements, especially if cross-domain child corpora are used.  This is extremely promising for overcoming the data scarcity problem which currently impedes progress in children’s speech recognition, particularly for non-native speakers.

Children’s speech recognition is a vast topic; thus, many ideas are left to be explored in future work, such as:
\begin{itemize}
    \item Designing a fine-tuning algorithm to handle the small vocal tracts of children,
    \item  Investigating augmentation methods beyond the ones mentioned in this paper,
    \item Including young speakers (such as Kindergarteners) in the pre-training data to improve age performance
    \item Studying the use of external Language Models, and
    \item Tackling short utterances when the pre-trained model is biased towards long utterances.
\end{itemize}
\bibliographystyle{elsarticle-num} 
\bibliography{refs}

\begin{thebibliography}{10}
\expandafter\ifx\csname url\endcsname\relax
  \def\url#1{\texttt{#1}}\fi
\expandafter\ifx\csname urlprefix\endcsname\relax\def\urlprefix{URL }\fi
\expandafter\ifx\csname href\endcsname\relax
  \def\href#1#2{#2} \def\path#1{#1}\fi

\bibitem{hair2018apraxia}
A.~Hair, P.~Monroe, B.~Ahmed, K.~J. Ballard, R.~Gutierrez-Osuna, Apraxia world:
  A speech therapy game for children with speech sound disorders, in:
  Proceedings of the 17th ACM Conference on Interaction Design and Children,
  2018, pp. 119--131.

\bibitem{mostow1994prototype}
J.~Mostow, S.~F. Roth, A.~G. Hauptmann, M.~Kane, A prototype reading coach that
  listens, in: AAAI, 1994, pp. 785--792.

\bibitem{russell1996applications}
M.~Russell, C.~Brown, A.~Skilling, R.~Series, J.~Wallace, B.~Bonham, P.~Barker,
  Applications of automatic speech recognition to speech and language
  development in young children, in: Proceeding of Fourth International
  Conference on Spoken Language Processing. ICSLP'96, Vol.~1, IEEE, 1996, pp.
  176--179.

\bibitem{ahmed2021auskidtalk}
B.~Ahmed, K.~Ballard, D.~Burnham, T.~Sirojan, H.~Mehmood, D.~Estival, E.~Baker,
  F.~Cox, J.~Arciuli, T.~Benders, et~al., Auskidtalk: an auditory-visual corpus
  of 3-to 12-year-old australian children's speech, in: Annual Conference of
  the International Speech Communication Association (22nd: 2021),
  International Speech Communication Association, 2021, pp. 3680--3684.

\bibitem{chung2019unsupervised}
Y.-A. Chung, W.-N. Hsu, H.~Tang, J.~Glass, An unsupervised autoregressive model
  for speech representation learning, arXiv preprint arXiv:1904.03240 (2019).

\bibitem{oord2018representation}
A.~v.~d. Oord, Y.~Li, O.~Vinyals, Representation learning with contrastive
  predictive coding, arXiv preprint arXiv:1807.03748 (2018).

\bibitem{ling2020deep}
S.~Ling, Y.~Liu, J.~Salazar, K.~Kirchhoff, Deep contextualized acoustic
  representations for semi-supervised speech recognition, in: ICASSP 2020-2020
  IEEE International Conference on Acoustics, Speech and Signal Processing
  (ICASSP), IEEE, 2020, pp. 6429--6433.

\bibitem{chung2020vector}
Y.-A. Chung, H.~Tang, J.~Glass, Vector-quantized autoregressive predictive
  coding, arXiv preprint arXiv:2005.08392 (2020).

\bibitem{vaswani2017attention}
A.~Vaswani, N.~Shazeer, N.~Parmar, J.~Uszkoreit, L.~Jones, A.~N. Gomez,
  {\L}.~Kaiser, I.~Polosukhin, Attention is all you need, Advances in neural
  information processing systems 30 (2017).

\bibitem{baevski2020wav2vec}
A.~Baevski, Y.~Zhou, A.~Mohamed, M.~Auli, wav2vec 2.0: A framework for
  self-supervised learning of speech representations, Advances in Neural
  Information Processing Systems 33 (2020) 12449--12460.

\bibitem{liu2020mockingjay}
A.~T. Liu, S.-w. Yang, P.-H. Chi, P.-c. Hsu, H.-y. Lee, Mockingjay:
  Unsupervised speech representation learning with deep bidirectional
  transformer encoders, in: ICASSP 2020-2020 IEEE International Conference on
  Acoustics, Speech and Signal Processing (ICASSP), IEEE, 2020, pp. 6419--6423.

\bibitem{song2019speech}
X.~Song, G.~Wang, Z.~Wu, Y.~Huang, D.~Su, D.~Yu, H.~Meng, Speech-xlnet:
  Unsupervised acoustic model pretraining for self-attention networks, arXiv
  preprint arXiv:1910.10387 (2019).

\bibitem{chi2021audio}
P.-H. Chi, P.-H. Chung, T.-H. Wu, C.-C. Hsieh, Y.-H. Chen, S.-W. Li, H.-y. Lee,
  Audio albert: A lite bert for self-supervised learning of audio
  representation, in: 2021 IEEE Spoken Language Technology Workshop (SLT),
  IEEE, 2021, pp. 344--350.

\bibitem{liu2021tera}
A.~T. Liu, S.-W. Li, H.-y. Lee, Tera: Self-supervised learning of transformer
  encoder representation for speech, IEEE/ACM Transactions on Audio, Speech,
  and Language Processing 29 (2021) 2351--2366.

\bibitem{wu2021transformer}
M.~Wu, K.~Li, W.-K. Leung, H.~Meng, Transformer based end-to-end
  mispronunciation detection and diagnosis., in: Interspeech, 2021, pp.
  3954--3958.

\bibitem{xu2021explore}
X.~Xu, Y.~Kang, S.~Cao, B.~Lin, L.~Ma, Explore wav2vec 2.0 for mispronunciation
  detection., in: Interspeech, 2021, pp. 4428--4432.

\bibitem{shibano2021speech}
T.~Shibano, X.~Zhang, M.~T. Li, H.~Cho, P.~Sullivan, M.~Abdul-Mageed, Speech
  technology for everyone: Automatic speech recognition for non-native english
  with transfer learning, arXiv preprint arXiv:2110.00678 (2021).

\bibitem{deng2021improving}
K.~Deng, S.~Cao, L.~Ma, Improving accent identification and accented speech
  recognition under a framework of self-supervised learning, arXiv preprint
  arXiv:2109.07349 (2021).

\bibitem{fan2021bi}
R.~Fan, A.~Afshan, A.~Alwan, Bi-apc: Bidirectional autoregressive predictive
  coding for unsupervised pre-training and its application to children’s asr,
  in: ICASSP 2021-2021 IEEE International Conference on Acoustics, Speech and
  Signal Processing (ICASSP), IEEE, 2021, pp. 7023--7027.

\bibitem{xu2021tal}
G.~Xu, S.~Yang, L.~Ma, C.~Li, Z.~Wu, The tal system for the interspeech2021
  shared task on automatic speech recognition for non-native childrens speech.,
  in: Interspeech, 2021, pp. 1294--1298.

\bibitem{ward2019my}
W.~Ward, R.~Cole, S.~Pradhan, My science tutor and the myst corpus (2019).

\bibitem{shobaki2000ogi}
K.~Shobaki, J.-P. Hosom, R.~Cole, The ogi kids’ speech corpus and
  recognizers, in: Proc. of ICSLP, 2000, pp. 564--567.

\bibitem{gretter2020tlt}
R.~Gretter, M.~Matassoni, S.~Bann{\`o}, D.~Falavigna, Tlt-school: a corpus of
  non native children speech, arXiv preprint arXiv:2001.08051 (2020).

\bibitem{gretter2020overview}
R.~Gretter, M.~Matassoni, G.~D. Falavigna, E.~Keelan, C.~W. Leong, Overview of
  the interspeech tlt2020 shared task onasr for non-native children’s speech,
  in: Interspeech 2020, 2020, pp. 245--249.

\bibitem{panayotov2015librispeech}
V.~Panayotov, G.~Chen, D.~Povey, S.~Khudanpur, Librispeech: an asr corpus based
  on public domain audio books, in: 2015 IEEE international conference on
  acoustics, speech and signal processing (ICASSP), IEEE, 2015, pp. 5206--5210.

\bibitem{graves2006connectionist}
A.~Graves, S.~Fern{\'a}ndez, F.~Gomez, J.~Schmidhuber, Connectionist temporal
  classification: labelling unsegmented sequence data with recurrent neural
  networks, in: Proceedings of the 23rd international conference on Machine
  learning, 2006, pp. 369--376.

\bibitem{shivakumar2022end}
P.~G. Shivakumar, S.~Narayanan, End-to-end neural systems for automatic
  children speech recognition: An empirical study, Computer Speech \& Language
  72 (2022) 101289.

\bibitem{kingma2014adam}
D.~P. Kingma, J.~Ba, Adam: A method for stochastic optimization, arXiv preprint
  arXiv:1412.6980 (2014).

\bibitem{hannun2014deep}
A.~Hannun, C.~Case, J.~Casper, B.~Catanzaro, G.~Diamos, E.~Elsen, R.~Prenger,
  S.~Satheesh, S.~Sengupta, A.~Coates, et~al., Deep speech: Scaling up
  end-to-end speech recognition, arXiv preprint arXiv:1412.5567 (2014).

\bibitem{knill2020non}
K.~M. Knill, L.~Wang, Y.~Wang, X.~Wu, M.~J. Gales, Non-native children's
  automatic speech recognition: The interspeech 2020 shared task alta systems.,
  in: INTERSPEECH, 2020, pp. 255--259.

\bibitem{katana}
\href{https://research.unsw.edu.au/katana}{Katana}.
\newline\urlprefix\url{https://research.unsw.edu.au/katana}

\bibitem{park2019specaugment}
D.~S. Park, W.~Chan, Y.~Zhang, C.-C. Chiu, B.~Zoph, E.~D. Cubuk, Q.~V. Le,
  Specaugment: A simple data augmentation method for automatic speech
  recognition, arXiv preprint arXiv:1904.08779 (2019).

\end{thebibliography}





\end{document}